# Technical Report: Competition Solution For Modelscope-Sora

Shengfu Chen, Hailong Liu and Wenzhao Wei

**Abstract.** This report presents the approach adopted in the Modelscope-Sora challenge, which focuses on fine-tuning data for video generation models. The challenge evaluates participants' ability to analyze, clean, and generate high-quality datasets for video-based text-to-video tasks under specific computational constraints. The provided methodology involves data processing techniques such as video description generation, filtering, and acceleration. This report outlines the procedures and tools utilized to enhance the quality of training data, ensuring improved performance in text-to-video generation models.

## 1 Introduction

The Modelscope-Sora challenge encourages participants to develop data-centric solutions for fine-tuning Sora-like models. These models are primarily used in text-to-video tasks, where the input text is transformed into a semantically coherent video output. High-quality data is critical for improving model performance in such tasks. The challenge involves training a model using LoRA (Low-Rank Adaptation) under two different resolution settings: 16x256x256 and 16x512x512, with the amount of training data restricted by pixel count limits. This report details the data processing pipeline, including video description generation, filtering, and video acceleration, to ensure the best possible training data is used.

## 2 Method

The video generation model combines visual and linguistic information to produce contextually meaningful video sequences. However, training an efficient video generation model relies on a large amount of high-quality data. Therefore, preprocessing, cleaning, and enhancing the raw video data become critical steps in ensuring the model's performance. The dataset for the Modelscope-Sora challenge consists of video data. The following sections detail the data processing methods, which mainly include video description generation, video filtering, and video acceleration. It is worth noting that the original dataset contains videos with existing textual descriptions as well as videos without them. Through testing, we found that using only the videos with existing textual descriptions and applying simple data processing yielded satisfactory results in the preliminary evaluation. Therefore, we applied different data processing workflows for videos with and without textual descriptions. Figure 1 shows the overall data processing flowchart.

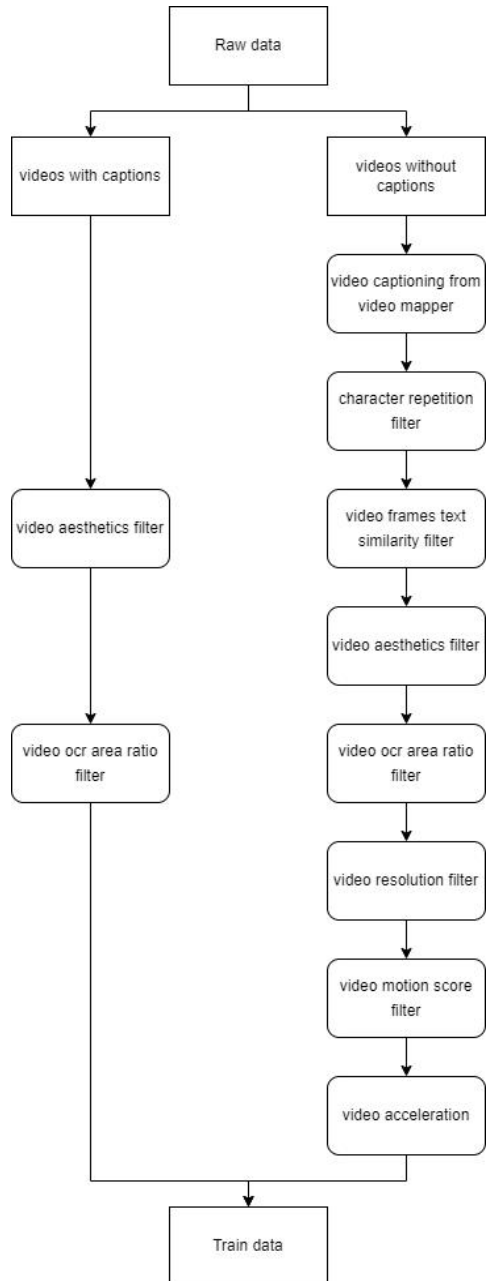

**Fig. 1.** Overall flowchart of data processing

## 2.1 Video Description Generation

In text-to-video generation tasks, the model takes text information as input into a Text-to-Video Model, generating target video outputs that are semantically consistent with the text content. Therefore, the video descriptions in the training data need to be not only detailed but also accurate, so that the model can learn the precise mapping between the text and video content. High-quality descriptions enhance the model's performance in understanding text and generating videos, ensuring that the generated video content is semantically coherent and visually consistent, meeting the diverse generation needs. In actual training, precise and detailed text descriptions help the model better capture complex features such as temporal information, scene transitions, and character actions in the video. This ensures that the generated videos not only align with the text content but also maintain high visual quality. Thus, optimizing the text descriptions in the training data is a key step in improving the performance of video generation models.

Typical video description generation algorithms include VideoBLIP[1], VideoChat2, and LLaVA[2], which excel in multimodal learning and video content understanding. After considering the performance of these algorithms (such as the accuracy of generated captions and contextual consistency) and efficiency (such as processing speed and resource consumption), we chose VideoBLIP as the video description generation tool to produce detailed and high-quality descriptions for the training data. VideoBLIP can generate high-quality, semantically rich descriptions, ensuring a high degree of consistency between the video description text and the video content, thus providing more effective training data support for our text-to-video generation model. We used the VideoBLIP model from data-juicer[3] to generate descriptions for the videos.

## 2.2 Video Filtering

To improve the generation quality of the video generation model, it is necessary to filter the videos and their corresponding text descriptions in the dataset. We used filtering operators provided by data-juicer to select high-quality training data. Referring to the data processing workflows from the Open-Sora and EasyAnimate[4] projects, and after experimental comparisons, we selected the following six effective video filtering methods.

### 2.2.1 Character Redundancy Filtering

During the process of generating video descriptions, there may be redundant or invalid data containing excessive character repetition. The purpose of character redundancy filtering is to remove these overly repetitive descriptions, ensuring the accuracy and readability of the text. Eliminating this redundant information helps prevent the model from learning noisy data, thereby improving the generation quality.

### 2.2.2  Video Frame and Text Similarity Filtering

To ensure that the generated subtitles are highly relevant to the video content, filtering the similarity between video frames and text is crucial. We use the CLIP model to calculate the similarity between video frames and generated subtitles. By calculating the similarity between frames and text, we ensure that only those samples with a high degree of match are retained. This approach filters out irrelevant or inaccurate subtitle descriptions by ensuring a visual-textual alignment between video frames and their descriptive subtitles. Pairs of frames and text with high similarity help the model better learn the relationship between video content and textual expression.

The core idea of CLIP is contrastive learning, which pulls together the embeddings of image-text pairs (i.e., images and texts describing the same scene) in the embedding space while pushing apart unrelated image-text pairs, thereby learning the semantic associations between images and text.

The training objective of CLIP is to maximize the similarity between correct image-text pairs while minimizing the similarity between incorrect image-text pairs, allowing the model to find matching images when given input text and vice versa. This approach enables a single model to process two different modalities of input: images and text.

### 2.2.3  Video Aesthetics Filtering

The aesthetic quality of videos is crucial in generating them, as visually appealing videos can enhance the output of the generation model. There are many low-quality videos in the dataset of this competition. We use an improved aesthetic physician to calculate the aesthetic score of video frames, remove those videos with poor visual quality, and ensure the high quality of training data. Figure 2 shows the statistical chart of video beauty credits in the dataset of this competition.

### 2.2.4  Video OCR Text Region Filtering

The video may contain a large amount of text information unrelated to the target content, such as subtitles: Many videos come with subtitles that may not be relevant to the video content and may even interfere with the model's understanding of the video image. Watermarks and Advertisements: Common watermarks or advertisements can appear in the corners or center of the screen of a video, affecting the overall visual perception of the video. The goal of training the text video task is to generate video content that is consistent with the input text. If there is too much textual information in the video, it may become noise, affecting the model's understanding of visual content and thus affecting the quality of the generated video. To avoid the interference of irrelevant text on model learning, we use OCR text region detection to identify and filter videos with excessive text content. Figure 3 shows the statistical chart of the OCR text area in the dataset of this competition.

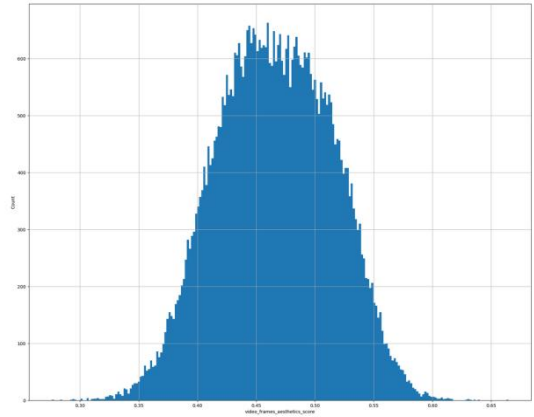

**Fig. 2.** Statistical Chart of Video Aesthetics Scores

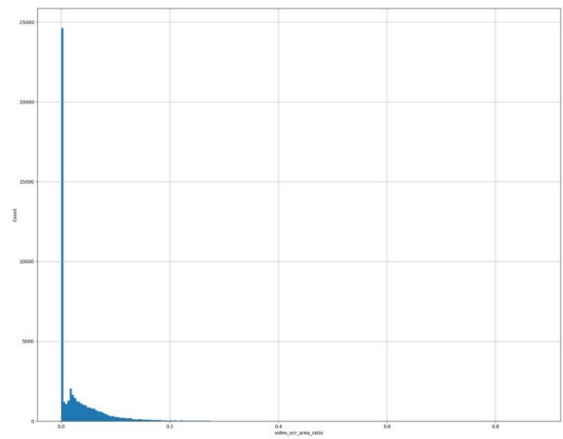

**Fig. 3.** Statistical Chart of Video OCR Text Region

### 2.2.5 Video Resolution Filtering

Low resolution videos can produce significant visual blur during training, which can make it difficult for the model to capture key visual details such as object edges, subtle movements, expressions, etc. Due to the need for the model to learn the detailed changes in different scenes during the training process, the blurring of low resolution videos can make it difficult for the model to generate clear video content, thereby reducing the quality of the final generated video. We obtain high-definition video data by filtering low resolution videos to improve the video generation quality of the model.

### 2.2.6 Video dynamic score filtering

Some videos may have minimal content changes and are almost static images, making it difficult for models to learn dynamic features corresponding to the text. These types of videos have low motion scores and cannot provide useful temporal information for literary video tasks. By filtering out these static videos, it can be ensured that the training set retains video samples with sufficient dynamic information. On the contrary, if the motion in the video is too intense or complex, and the motion score is too high, it may introduce noise, making it difficult for the model to capture the accurate correspondence between text and video content. For certain tasks, videos with excessive physical activity may affect the model's ability to generate clear and natural dynamic videos. By filtering out such videos, unnecessary complex actions can be avoided in generating videos. By filtering through video exercise scores, we can select videos with moderate exercise amplitude, avoiding videos in the training set that are too static or have intense exercise. Figure 4 shows the statistical chart of video sports scores in the dataset of this competition.

### 2.3 Video Filtering

In the process of analyzing the dataset, we found a large number of slow playback and high-quality high-resolution videos. In order to better utilize these training data, we used ffmpeg to perform video acceleration processing on these videos to enhance their sense of motion. In our testing, this approach significantly improved the dynamic score in the evaluation metrics.

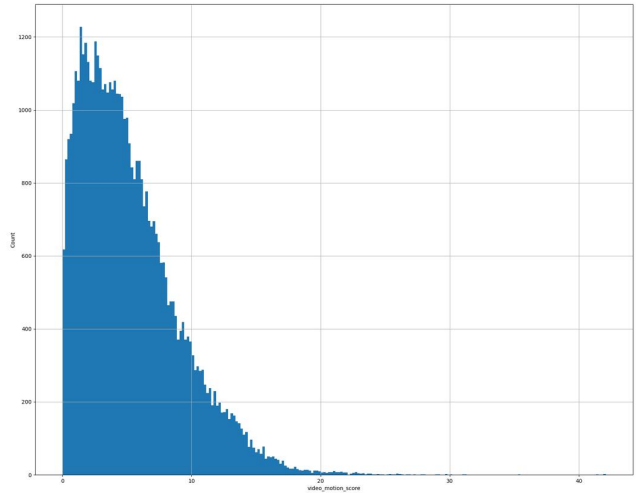

**Fig. 4.** Video dynamic score statistics chart

## 3      Conclusions

In this report, we have detailed the data processing techniques applied during the Modelscope-Sora challenge to improve the performance of text-to-video generation models. Key methods included video description generation, filtering of redundant and low-quality data, and video acceleration. By employing these techniques, we were able to ensure high-quality training data, which directly contributed to the model's ability to generate semantically coherent and visually appealing videos. The use of filtering methods like character redundancy removal, frame-text similarity filtering, and video aesthetics evaluation helped eliminate noise and irrelevant information from the dataset. Moreover, video acceleration enhanced the training data's dynamic content, further improving the model's temporal understanding. These approaches resulted in improved overall model performance, especially in generating high-quality, contextually relevant video outputs. Future efforts could explore additional optimization techniques to further enhance both the data processing pipeline and the efficiency of training large-scale video generation models.